# Automated Triaging and Transfer Learning of Incident Learning Safety Reports Using Large Language Representational Models

**Short Running Title: Incident Report Triage Using Machine Learning**

Peter Beidler, MAB MD[1*], Mark Nguyen, MD[1*], Kevin Lybarger, PhD[2], Ola Holmberg, PhD[3], Eric Ford, PhD[4], John Kang, MD PhD[4]

\* These authors contributed equally to this research and share first authorship

1 University of Washington School of Medicine

2 University of Washington, Department of Biomedical Informatics and Medical Education

3 International Atomic Energy Agency, Radiation Protection of Patients Unit

4 University of Washington School of Medicine, Department of Radiation Oncology

**Corresponding Author:** John Kang (jkang3@uw.edu)

**Author Responsible for Statistical Analysis:** Mark Nguyen (son94@uw.edu)

**Conflict of Interest**: None

**Funding:** None

**Data Availability Statement:** Research data are stored in an institutional repository and will be shared upon request to the corresponding author.

**Abstract**

**PURPOSE:**

Incident reports are an important tool for safety and quality improvement in healthcare, but manual review is time-consuming and requires subject matter expertise. Here we present a natural language processing (NLP) screening tool to detect high-severity incident reports in radiation oncology across two institutions.

**METHODS AND MATERIALS:**

We used two text datasets to train and evaluate our NLP models: 7,094 reports from our institution (Inst.), and 571 from IAEA SAFRON (SF), all of which had severity scores labeled by clinical content experts. Each dataset was split into train, validation, and test sets. We trained and evaluated two types of models: baseline support vector machines (SVM) and BlueBERT which is a large language representation model pretrained on PubMed abstracts and hospitalized patient data. We assessed for generalizability of our model in two ways. First, we evaluated models trained using Inst.-train on SF-test. Second, we trained a $BlueBERT_{TRANSFER}$ model that was first fine-tuned on Inst.-train then on SF-train before testing on SF-test set. To further analyze model performance, we also examined a subset of 59 reports from our Inst. dataset, which were manually edited for clarity.

**RESULTS**

Classification performance on the Inst. test achieved AUROC 0.82 using SVM and 0.81 using BlueBERT. Without cross-institution transfer learning, performance on the SF test was limited to an AUROC of 0.42 using SVM and 0.56 using BlueBERT. $BlueBERT_{TRANSFER}$, which was fine-tuned on both Inst.-train and SF-train sets, improved the performance on SF test to AUROC 0.78. Performance of SVM, and $BlueBERT_{TRANSFER}$ models on the manually curated Inst. reports (AUROC of 0.85 and 0.74) was similar to human performance (AUROC of 0.81).

**CONCLUSION:**

In summary, we successfully developed cross-institution NLP models on incident report text from radiation oncology centers. These models were able to detect high-severity reports similarly to humans on a curated dataset.

**Main body**

**Introduction**

Incident learning systems (ILS) have become a major tool for reducing medical errors and are the basis for safety improvement in radiation oncology[1–5]. These systems consist of voluntary or mandatory reporting systems for errors, often including near-misses, and systematic analysis of reports to guide quality improvement. Reports generally consist of free-text incident descriptions and can be submitted by physicians, physicists, nurses, radiation therapists, or any other professionals involved in patient care.

Ford et al.[2] suggested a framework (**Figure 1**), for departmental incident learning systems to include initial triage for investigation and possible immediate action followed by either immediate root-cause analysis or presentation to a quality assurance committee at regular meetings for further review. Timely and accurate initial triage requires someone with extensive experience both in the ILS and in all areas of patient care workflow. Mullen et al[6] reported bias in incident severity scoring based on profession and involvement in Quality Assurance (QA) and Quality Improvement (QI) meetings. They reported "fair agreement" among 28 employees involved in patient care at a radiation oncology department with a Krippendorff's alpha score of 0.376. Initial event triaging is both resource-intensive and prone to disagreement even among the most experienced raters. Given the existence of many large repositories of incident reports, machine learning (ML) tools have potential to improve the consistency and efficiency of initial triage. This need has been described by Syed et al., Wang et al., and Ford et al.[2,7,8] (**Supplementary Text S1**)

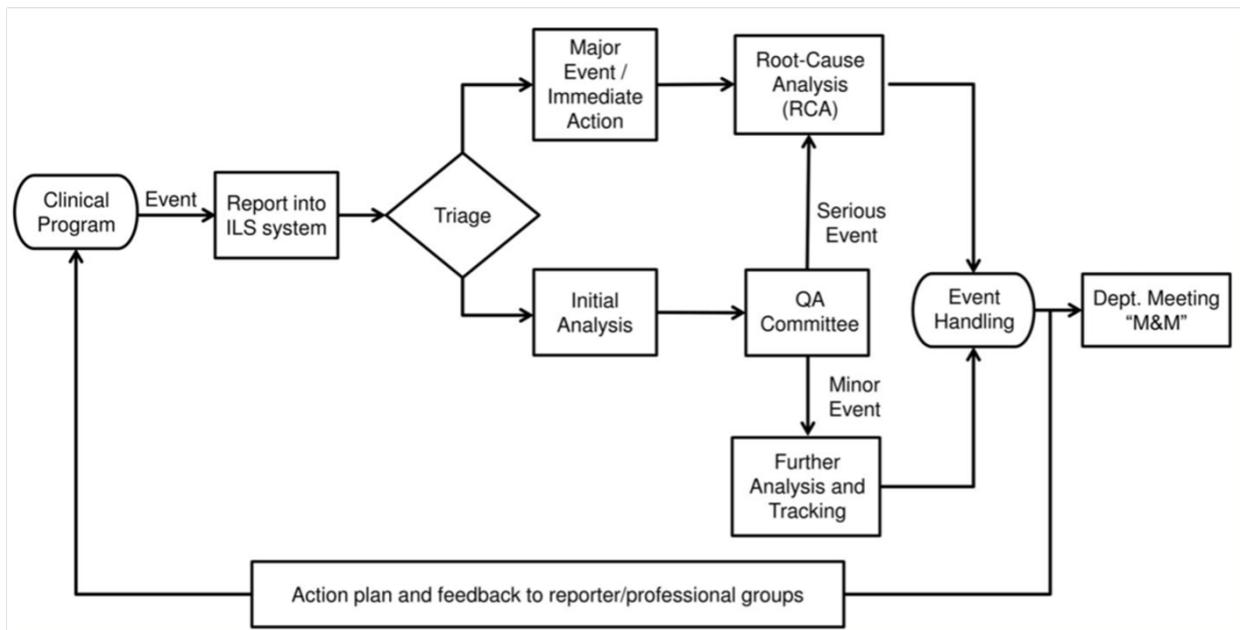

**Figure 1**: Schematic representation of the operation of an institutional incident learning system. Used with permission from the authors of Ford et al. 2018.

The development of a generalized ML classifier that can be deployed across institutions will be challenging for several reasons. Many different severity classification systems have been developed, and none are standardized[1,9–13]. Each of these scoring systems weigh actual event severity, possible event severity, and likelihood of actualization differently. Thus, the same report text could have different severity scores at two institutions. Additionally, report content varies widely because of different workflows, equipment, and safety culture. National and international repositories such as Safety in Radiation Oncology (SAFRON)[14], Radiation Oncology Incident Learning System (RO-ILS)[15], and European Society for Radiotherapy and Oncology (ESTRO)[16] may be more representative of the full range of safety report topics in radiation oncology.

Because of these differences between institutions and the extra-legal protections afforded to incident reports in the United States against discovery, it is unlikely that one machine learning model could be deployed across institutions. With recent advances in large language models (LLMs) and transfer learning, we hypothesize that models could be developed using larger datasets and fine-tuned on smaller datasets for better performance. Transfer learning involves adapting a model initially developed for one task to a different task, leveraging the knowledge gained during the initial training to reduce the time and data needed for effective learning on the new task. In this study, we trained and evaluated two machine learning models using two distinct datasets: a large departmental repository of incident reports from our institution (Inst.) and the SAFRON dataset (SF), a smaller though more highly curated repository representing over 50 registered medical facilities and hospitals around the world. To evaluate cross-institutional generalizability, we systematically assess external validity by training on one dataset and testing on the other.

**Method and materials**

Models were trained and evaluated on the Inst. and SF datasets described below. Dataset size and report lengths are described in **Table 1**. The distribution of severity labels of reports in the Inst. and SAFRON datasets are shown in **Figure 2a & b**.

**Table 1**: Datasets used to train and evaluate ML models. Inst. refers to our institution's dataset. SF refers to SAFRON dataset.

| Dataset | Number of Reports | Report Length (words, median ± std) | High-severity (%) |
|---|---|---|---|
| Inst. | 7,094 | 44 ± 41 | 15.6 |
| SF | 1,684 (571 labeled) | 71 ± 69 | 34.2 |

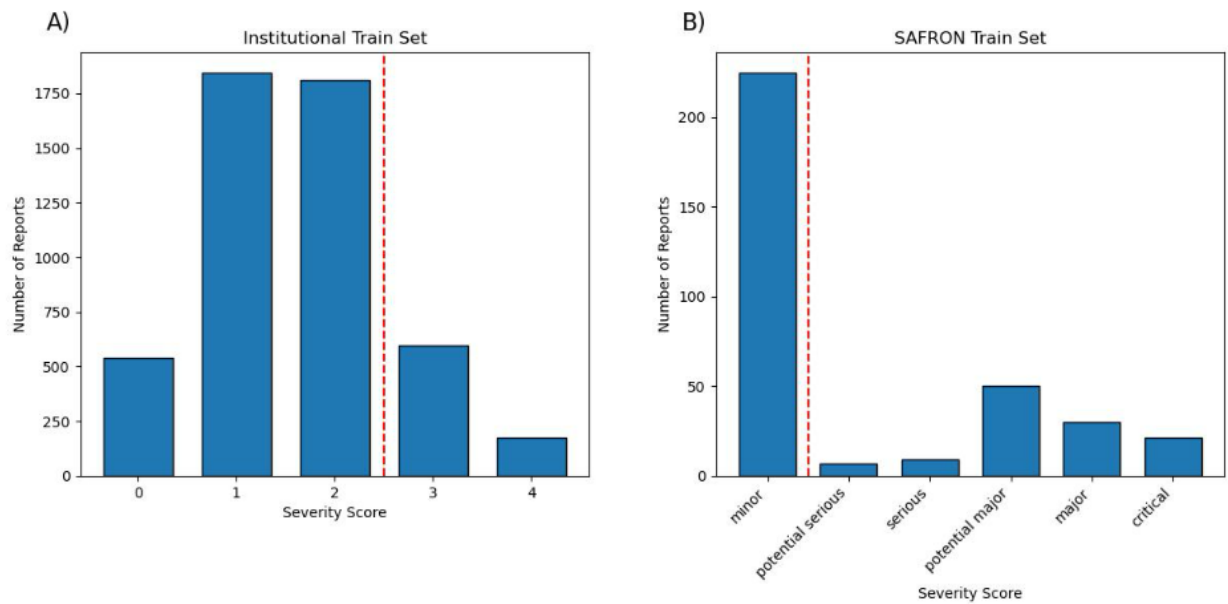

**Figure 2:** Severity score distributions in the (A) Inst. and (B) SAFRON train sets. The Inst. dataset uses a numerical severity scale ranging from 0 to 4, while the SAFRON dataset applies six categorical severity levels to describe incident reports: minor, potential serious, serious, potential major, major, and critical. In this study, we binarized severity such that Institutional reports with severity scores less than 3 were classified as "low-severity," and those with scores of 3 or 4 were classified as "high-severity." For the SAFRON dataset, reports labeled "minor" were considered "low-severity," while all other categories were defined as "high-severity."

Institutional Dataset

The Institutional dataset (Inst.) consisted of 7,094 free text reports collected from 2012 through 2020 from a Radiation Oncology departmental incident learning system described by Nyflot et al. and Kusano et al.[9,17]. Each report is assigned a severity score using the near-miss risk index (NMRI) by a weekly interdisciplinary panel of physicists, dosimetrists, physicians, and radiation therapists. The NMRI scoring system consists of integer values from 0 to 4. We consider scores of 3 and 4 to be high-severity and 0 through 2 to be low-severity. Topics of reports include errors in medical therapy, software issues, and miscommunication. In addition, after the initial report is submitted, further text can be added to the report as more information is gained to provide a more nuanced description of the event. Examples of an Inst. report are shown in **Supplementary Table S1**.

SAFRON Dataset

The SAFRON dataset (SF) consisted of 1,684 reports, of which 571 had associated ordinal severity labels chosen from the following options: "minor", "potential serious", "serious",

"potential major", "major", "critical"[14]. After manual review by a safety committee member at our institution, SF reports assigned as "minor" were considered low-severity, and all other grades were considered high-severity, as this most closely mimicked the Inst. scoring system. These reports are much more heterogeneous than the Inst. set, as they originate from institutions across several different countries. Examples of a SF report are also shown in **Supplementary Table S1**.

Machine learning models

We implemented two types of machine learning models: Support Vector Machine (SVM) with TF*IDF encoding and BlueBERT[18], both described below. These models are both intended to be used for classification tasks.

Support vector machines, use a mapping function and decision boundary to assign inputs as one of two classes[19,20] The SVMs were implemented using the Sci-Kit Learn SVC function with default parameters. The regularization parameter C was tuned to optimize F1 score on Inst. and SF validation sets.

BlueBERT consists of the Bidirectional Encoder Representation Transformer large language model with ~110 million parameters that has been pretrained on more than 4 billion words collected from PubMed Open Access and more than 400 million words from MIMIC-III clinical text of hospitalized patients[18]. BlueBERT has an encoder that takes in text directly and does consider word order. The pretrained model weights were downloaded from Huggingface Version 4.26 using the AutoModel and AutoTokenizer classes.

Preprocessing

Both Inst. and SF datasets were preprocessed before model training Acronyms were expanded using a manually created dictionary specific to the domain, with an average of 1.22 expansions per report (**Supplementary Table S2**). All text was lower-cased. The Inst. and SF sets were randomly divided into train, validation, and test sets using 70/15/15 train/validation/test split for Inst. dataset and 60/20/20 split for the SF dataset.

Training procedure

We trained the SVM models using the term frequency-inverse document frequency (TF-IDF) representations of incident reports as input. TF-IDF, a non-semantic statistical measure of words' importance in a set of documents, was computed using Scikit-learn (Version 1.2) with English stop words removed and a minimum document frequency threshold of 10.

We also fine-tuned two distinct BlueBERT models: $BlueBERT_{Inst.}$, and $BlueBERT_{TRANSFER}$. The $BlueBERT_{Inst.}$ was fine-tuned using the Inst. dataset only. $BlueBERT_{TRANSFER}$ employs cross-institutional transfer learning by being first fine-tuned on the Inst. train set and subsequently on the SF train set.

For both BlueBERT models, a randomly initialized feed-forward classification layer was appended with one output dimension with sigmoid activation. Both were fine-tuned using a binary cross entropy loss with Adam optimizer and learning rates of $10^{-6}$ for Inst. and $10^{-8}$ for SF train set. Loss was weighted by train set inverse frequencies. Reports were truncated at 150 word-tokens for both models (95.6th and 87.7th percentile token length for Inst. and SAFRON train set reports, respectively, **Supplementary Figure S1**) due to memory constraints and batch size set to 32. The learning rate hyperparameter was tuned using validation loss curves and F1 score. Each of the models was fine-tuned 5 times with different random initializations of the classification layer. The model with the best F1 score was then used for further analysis.

For all models, validation sets were used to tune hyperparameters, and evaluating performance on test sets was only performed once model development was completed. Models were trained using an AMD Ryzen 7 5800X CPU and a single NVIDIA GeForce 3070Ti with 8GB VRAM with a training time of approximately 8-10 minutes.

Evaluation

We report classification performance using area under the Receiver Operating Characteristic (AUROC). The AUROC score on the Inst. and SF validation sets was used to select the best model for each dataset to further examine the performance on the test sets. While model's AUROC provides a useful summary of overall performance, decision thresholds must be defined for practical implementation. To demonstrate the tunability of our models at specific organizations, we also report performance at thresholds with alert rates of 20% and 50%, where alert rate is defined as the number of cases classified as high risk divided by the total number of cases. Adjusting the alert rate allows organizations to account for alert fatigue and balance sensitivity and specificity[21]. For example, increasing the alert rate increases true-positive detection (higher sensitivity) at the cost of increasing false positives (lower specificity).

Model versus human comparison

Since 59 reports of the Inst. dataset were previously curated and written in more natural language for a study on inter-rate agreement by Mullen et al. at our institution[6], we decided to selectively compare model performance on these reports with human performance. These reports were previously manually selected and edited for clarity for a different study in which 26 radiation oncology department employees assigned severity scores using the same 0-4 Inst. scale to each report to investigate inter-rater agreement. Each individual rater score was treated as a single prediction, resulting in a total of 709 human predictions used to calculate overall human performance. To avoid data leakage, these reports were also excluded from the Inst. dataset used for model training. Three examples of unedited and edited reports are shown in **Supplementary Table S3** to demonstrate levels of text clarity. The 59 reports used in this prior analysis were then fed to our SVM, BlueBERT$_{Inst.}$ and BlueBERT$_{TRANSFER}$ models. The model's AUROC using these reports was compared with the rater AUROC for this model versus human comparison study. For AUROC calculation of human raters, the ground truth was considered to be the consensus committee score.

**Results**

General Model Performance

The transfer learning BlueBERT$_{TRANSFER}$ had the best performance to detect high-severity reports across institutions. On the SF test set, the best BlueBERT$_{TRANSFER}$ model achieved an AUROC of 0.78, outperforming both the best SVM model (AUROC 0.42) and the best BlueBERT$_{Inst.}$ model (AUROC 0.56) (**Figure 3a**). In contrast, on the Inst. test set, there is no significant difference in model performance among SVM (AUROC 0.82), BlueBERT$_{Inst.}$(AUROC 0.81), and BlueBERT$_{TRANSFER}$ (AUROC 0.79) (**Figure 3b**).

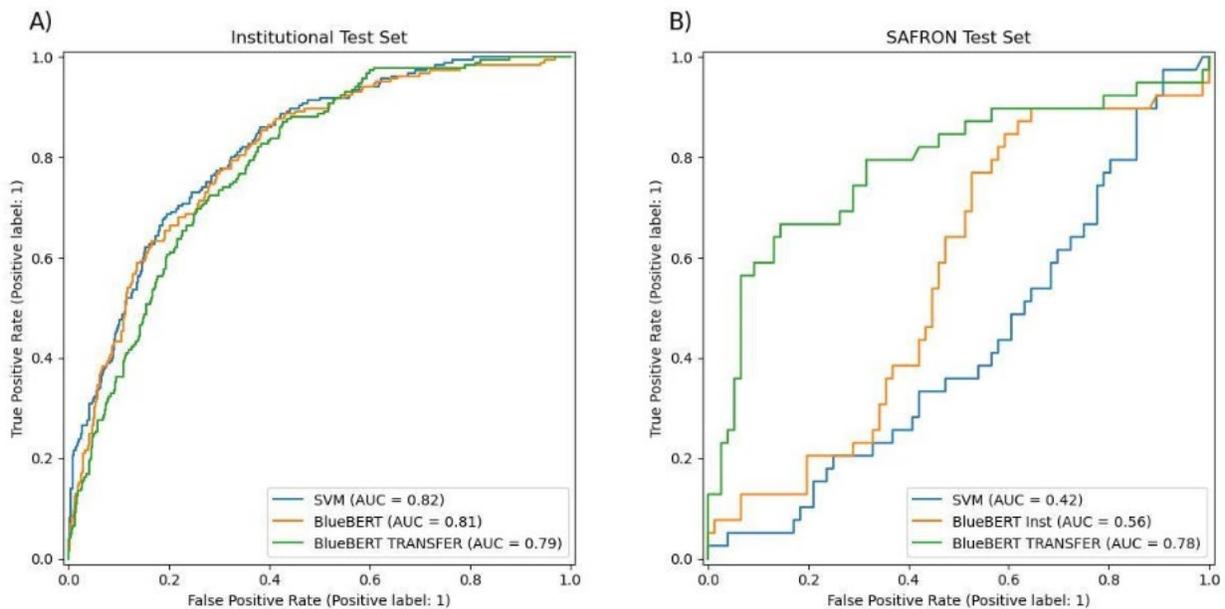

**Figure 3**: Receiver Operating Characteristic (ROC) curves for machine learning models evaluated on the Institutional Test Set and (B) the SAFRON Test Set. AUROC values are shown in the legend for each model.

Decision thresholds at a specific alert rate

While the model's AUROC provides a useful summary of overall performance, alert rate—and alert fatigue—can play a more significant role than performance with regards to clinical adoption. When evaluated on the SF Test Set, the BlueBERT$_{TRANSFER}$ model has a sensitivity of 0.46 and specificity of 0.93 with an alert rate set to 20%; and a sensitivity of 0.79 and specificity of 0.64 when alert rate was set at 50% (**Table 2** and **Table 3**). In other words, when increasing the alert rate from 20% to 50%, the BlueBERT$_{TRANSFER}$ model's true-positives increase by 70% (18 to 31 cases) while false-positive predictions increase more than 5-fold (5 to 27 cases).

**Table 2**: Classification performance of BlueBERT$_{TRANSFER}$ model on SF test set when alert rate was set to 20%. Sn = sensitivity; Sp = specificity; PPV = positive predictive value; NPV = negative predictive value.

|  |  | Label | | |
| --- | --- | --- | --- | --- |
|  |  | High-severity | Low-severity | |
| Prediction | High-severity | 18 | 5 | PPV = 0.78 |
| | Low-severity | 21 | 71 | NPV = 0.77 |
|  |  | Sn = 0.46 | Sp = 0.93 | |

**Table 3:** Classification performance of BlueBERT$_{TRANSFER}$ model on SF test set when alert rate was set to 50%. Sn = sensitivity; Sp = specificity; PPV = positive predictive value; NPV = negative predictive value.

|  |  | Label | | |
| --- | --- | --- | --- | --- |
|  |  | High-severity | Low-severity | |
| Prediction | High-severity | 31 | 27 | PPV = 0.53 |
| | Low-severity | 8 | 49 | NPV = 0.86 |
|  |  | Sn = 0.79 | Sp = 0.64 | |

Model versus human comparison results

For the 59 curated institutional reports, although SVM (AUROC 0.85) outperformed BlueBERT$_{Inst.}$ (AUROC 0.76) and BlueBERT$_{TRANSFER}$ (AUROC 0.74), these models demonstrated performance comparable to human raters (AUROC 0.81).

**Discussion**

We developed and validated machine learning models for severity screening of operational safety reports from a large Inst. Dataset, and evaluated external validity using a smaller, multi-national dataset (SAFRON). Our transfer learning approach BlueBERT$_{TRANSFER}$, which sequentially utilized both datasets, achieved an AUROC of 0.78 on the SF test set (**Figure 3**). This result not only shows the generalizability of our model but also highlights the potential of fine-tuning large language models to adapt across institutions. Additionally, our models (SVM, AUROC 0.85;

BlueBERT$_{Inst.}$, AUROC 0.76; and BlueBERT$_{TRANSFER}$, AUROC 0.74) have comparable performance when compared to individual department employees (AUROC 0.81).

Transfer learning

The BlueBERT$_{TRANSFER}$ model, which was fine-tuned sequentially on the Inst. and SF train set, achieved a substantial improvement (AUROC 0.78) compared to both the performance of standalone BlueBERT$_{Inst.}$ model (AUROC 0.56) and SVM (AUROC 0.42) on the SF test set (**Figure 3**). This performance boost shows that LLMs could be used to adapt to datasets with different institutions' linguistic, cultural, or procedural characteristics. Our results suggest that pretrained language models, when fine-tuned on small amount of local data, can serve as adaptable triaging tools. For reference, the SAFRON dataset contained only 571 reports compared to 7,094 in the Inst. dataset, which would make training a model from scratch challenging because of limited data. Through transfer learning the best BlueBERT$_{TRANSFER}$ model had a good cross-institutional triaging performance, correctly identifying 31 out of 39 high-severity reports and 49 out of 76 low-severity reports on the SF test set at an alert rate of 50%.

Model Performance

Without cross-institution transfer learning, BlueBERT underperformed compared to SVM on the Inst. test set. Great effort was made to optimize training of the BlueBERT models, including extensive hyperparameter tuning, further masked language modeling pretraining, and multi-task training using both the Inst. and SF datasets (**Supplementary Text S2**). Other external validity studies (**Supplementary Text S3** and **Supplementary Table S4**) are also included in the **Supplement**.

Several factors could contribute to the weaker performance of the BlueBERT model. Firstly, the Inst. reports contain highly sensitive text with domain-specific terminology that may not have been well-represented in BlueBERT's pretraining corpus (PubMed and clinical text). Moreover, label inconsistency of the reports could also limit our model performance. Notably, labels of the Inst. dataset are inconsistent with only a "fair" inter-rater agreement with Krippendorff's α of 0.376 in prior studies[6]. Labeling variability is likely due in part to the nine-year span of report collection, during which staff turnover, changes in equipment, and evolving workflows introduced new terminology and incident types. Additionally, many reports lacked sufficient clinical detail, which may have limited a more complex model's ability to accurately learn and assess severity. For example, a report reads: "Wrong headrest used for one fraction of patient treatment, leading to suboptimal patient positioning." Without knowing the anatomical site of treatment and the total number of fractions, it is unclear whether this report should be labeled 2 (low-severity) or 3 (high-severity).

Despite these challenges, our results support the value of transfer learning. The benefit of transfer learning, as described above as well as by Syed et al.[7], suggests that it may be useful for developing severity screening tools on other datasets as well, even if they use different scoring systems and the nature of the reports is quite different (as shown in **Figure 2** and **Supplementary Table S2**). As a result, while SVM models showed marginally better results on the Inst. dataset, the adaptability of the BlueBERT models allows other institutions to implement a severity triaging model and achieve robust performance with additional fine-tuning with only a small dataset.

While we report model performance at alert rates of 20% and 50% for the sake of demonstration, it is critical to discuss with clinical representatives what an acceptable alert rate would be at each institution.

**Conclusion**

We successfully developed machine learning models for automated severity screening of radiation oncology incident reports. Transfer learning with BlueBERT enhances generalizability and significantly improves performance on an external and much smaller dataset. Additionally, our work demonstrated that ML has the potential to standardize and streamline incident triaging, as we could achieve robust performance similar to human evaluators. These findings highlight the potential of NLP to improve efficiency and consistency in safety reporting systems, allowing for rapid triaging of incident reports for immediate intervention.

**Supplemental materials**

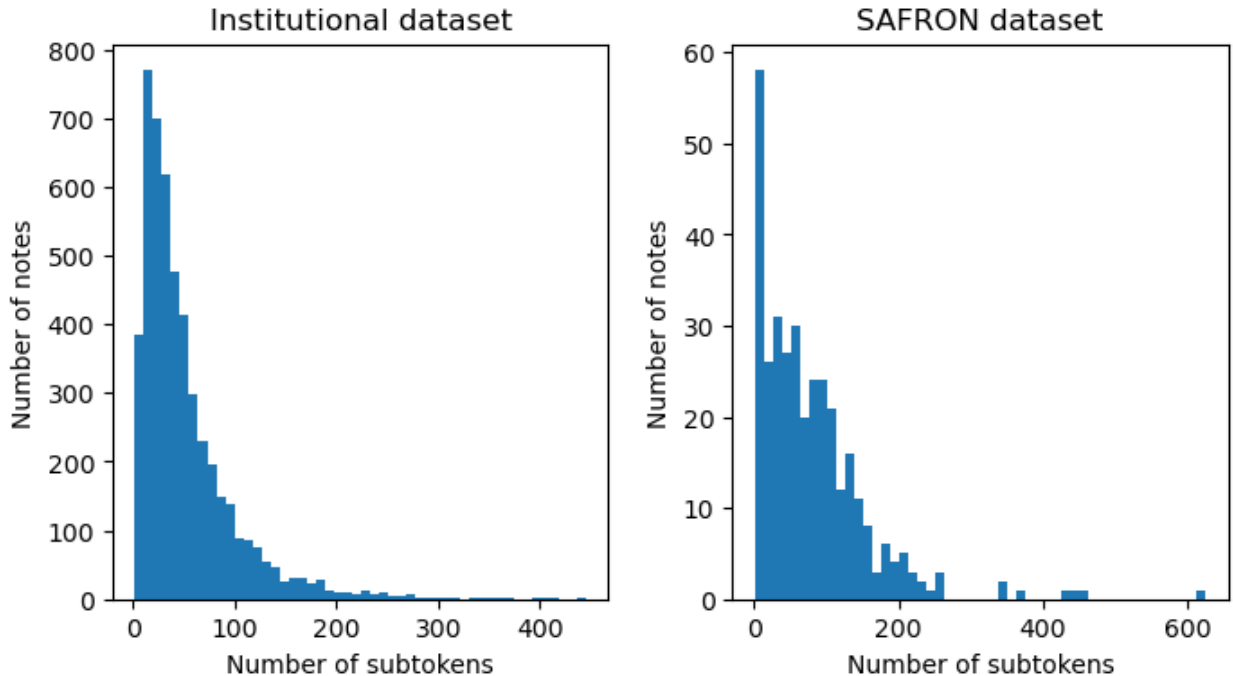

**Supplementary Figure S1**: Lengths of reports after BlueBERT model tokenization

**Supplementary Text S1:** Similar Work

To our knowledge, only two other studies have attempted to automate severity classification of operational safety reports in radiation oncology. These studies are built on relatively small datasets and use poorly representative metrics of model performance.

Syed et al.[7] used ULMFiT which is an early transfer learning model to predict severity of incident reports using two small datasets totaling 875 reports from the Virginia Commonwealth University and Veterans Health Administration and reported F1 scores of 0.81 and 0.68. While these scores are higher than F1 scores that we report on the Institutional and SAFRON datasets, it should be noted that the authors reported macro-averaged F1 scores instead of micro-averaged, which is more common for binary classification. Macro-averaged F1 tends to be higher than micro-averaged if the positive class (high-severity) is rare, as is the case in their VCA and VHA datasets. Additionally, their study did not address that report word-length appeared to correlate with report severity.

Another study by Wang et al.[8] developed ensembled SVM and logistic regression classifiers to predict severity scores of radiation oncology incident reports from Australia in a one-versus-all scheme. Their models were trained on 1,160 reports and reported macro-averaged F1 scores of 0.501 and 0.527 on unbalanced test sets from the training database and from a separate report database, respectively. Although they demonstrated direct generalizability to a separate dataset with similar F1 score, it is unclear how similar the report content is between datasets, and severity scoring was done using the same scoring system with high inter-rater agreement. Furthermore, their two testing datasets were highly unbalanced with only 0.52% and 0.38% of reports being the highest severity class. A key advantage of this work was that their training datasets were de-identified which could enable model sharing. However SVM and logistic regression are less amenable to transfer learning than modern transformer models.

These studies demonstrate a recent effort to partially automate the notoriously difficult task of interpreting incident reports in radiation oncology. In contrast to their approaches, our SVM model is built on a larger dataset while the BlueBERT transfer learning model demonstrates transfer learning to a notably different, diverse, international dataset.

**Supplementary Table S1**: Representative examples of reports from Institutional (left column) and SAFRON (right column) datasets. Individuals' initials are anonymized.

| **Institutional Report Example** | **SAFRON Report Example** |
|---|---|
| No new PDF in patient plan for Iso marking. DRR from 1252013 was brought up in CT sim room and VISUALLY matched old accidentally printed PDF from 122209. 2009 coordinates were 1.85 - 3.39 - 1.15 vs 2013 coordinates of 2 - 3.5 - 3. XX was called. | Patient has tattoos to mark the isocenter position. According to our process organization, we perform in this case daily setup images to correct the positioning of the patient. This had not been prepared by the planning team. During the first treatment field of the 2nd fraction, this mistake was realized and corrected. |
| ABC patient- at sim her breath hold was on EXHALE. In her ABC file, it was marked as INHALE. She wasnt able to complete her vsim and said what we were doing was too different. The next day, XX went over the whole process with her again and discovered the error. She was able to complete her vsim after that. Can we make a note in NavTab of how an ABC patient is simmed Add it to SBRT setup sheet or checklist TR | Patient has pacemaker, and it was decided that during irradiation the RTTs should place a magnet above the device. This was correctly noted in the setup notes and in the chart, conform to our standard. However, as physics chart check was delayed, the patient was positioned without mode up of the treatment plan, but only with the printout of the plan at hand. During mode up later at the console, the setup note is not immediately visible and therefore the information about the magnet did not reach the treatment team. |
| Patient had TBI schedule changed from 6/30-7/2 2Gy x 6 to 6/28-7/1 1.65 x 8 Dose and fractionation changed less than 2 business days before TBI without transplant team telling us; fortunately caught by MD when reviewing CCOs. But calcs had to be redone urgently | An institution did not own a barometer and the physicist relied on the local airport for measurement of atmospheric pressure. Pressure reported by the airport was corrected to sea level, but the physicist interpreted the pressure report as appropriate for the elevation of the site, about 1000 m. The use of the incorrect pressure resulted in errors in the measurement of machine outputs, which caused a 13% overdose to patients. The error affected the calibration of all machines at this institution. INITIATING EVENT: Calibration error: The use of incorrect pressure values to correct for atmospheric pressure resulted in an incorrect beam output. (IAEA Safety Reports Series No. 17, Event No. 3) |

**Supplementary Table S2**: Glossary of Abbreviations and Expansions

| Abbreviation | Expansions |
|---|---|
| 'pt' | 'patient' |
| 'pts' | 'patients' |
| 'sim' | 'simulation' |
| 'sbrt' | 'stereotactic body radiation therapy' |
| 'ssd' | 'source to surface distance' |
| 'rx' | 'prescription' |
| 'tx' | 'treatment' |
| 'imrt' | 'intensity-modulated radiation therapy' |
| 'appt' | 'appointment' |
| 'rtp' | 'radiation treatment planning' |
| 'ro' | 'radiation oncology' |
| 'qcl' | 'quality check lists' |
| 'h p' | 'history and physical' |
| 'mq' | 'mosaiq' |
| 'dosi' | 'dosimetry' |
| 'tbi' | 'total body irradiation' |
| 'drr' | 'digitally reconstructed radiograph' |

**Supplementary Table S3**: Selected original Inst. reports and manually edited report for clarity for inter-rater study. These reports were adopted for ML versus human performance comparison.

| Original Inst. report | Edited report for Mullen et al. (adopted for our comparison study) | Severity Score (Original) |
|---|---|---|
| l scf neck boost field was planned for 5 fx instead of 3 fx as prescribed.planned for incorrect number of fx ? good catch by therapists [track] compressed timeframe < 24hrs physics check not done prior to therapist audit. | Left supraclavicular/neck boost field was planned for 5 fractions instead of 3 fractions as prescribed. Comments: Good catch by therapists before patient arrived for filming. Compressed timeframe < 24hrs. Physics had not yet done check. | 3 |
| physics approved e field with no mus in itan e- fld was approved by physics in di with no mus entered. mw | Physics approved an electron field with no MUs in it Comments: An electron field was approved by physics in DI with no MUs entered. Therapists alerted him to fix. Rush job. Therapists noted no work done until last minute (QCL 'got lost'). | 0 |
| pod did not respond to stereotactic body radiation therapy call.no answer on the pod phone x3 and 3 physicists paged. no response. we proceeded with the stereotactic body radiation therapy pre scan. db | Physicist of the day did not respond to SBRT call. Comments: No answer on the POD phone x3 and 3 physicists paged. No response. We proceeded with the SBRT pre scan. | 1 |

**Supplementary Text S2**: Mask language model and multitask training methods:

To thoroughly evaluate the potential of NLP in analyzing multi-center ILS database, we incorporate 2 additional state-of-the-art BlueBERT training methods, including masked language model and multitask training. However, there was no significant improvement in model performance.

Multitask training (MT) is a learning approach where a single model is trained on multiple related tasks simultaneously, compared to learning the tasks independently as in the BlueBERT$_{TRANSFER}$ training procedure. This method leverages the commonalities across tasks to provide more data for the model to learn from, which can lead to more robust and generalized learning. Multitask training was performed by alternating batches of SAFRON and Institutional training sets, with two different classifier layers and learning rate of 1e-6. One epoch was defined as one pass through the Institutional train set with batches of 32 (16 from each set).

Masked language model (MLM) training is a technique used primarily in the pre-training of language models like BERT. In this approach, some percentages of the input tokens are randomly masked or hidden from the model during training. The model's objective is then to predict the original identity of the masked words, given their context within the sentence. This process teaches the model to understand and predict words based on their context. We used this approach to further fine-tune the BlueBERT model using the institutional training set, SAFRON training set, and SAFRON reports with no associated severity score, totaling 266,911 words. This was performed using the HuggingFace AutoModelML class with parameters described in supplement.

**Supplementary Text S3**: Evaluating external validity using Nyflot and RO-ILS datasets.

To better understand the external validity of our ML models, we performed inference on 13 additional reports from RO-ILS case reports and 11 artificial reports published by Nyflot et al.[9] to describe the NMRI scoring system of our Inst. dataset. The average word count of the RO-ILS and Nyflot dataset are 118 ± 78 and 26 ± 8, respectively. The models were evaluated using F1 and ROC AUC metrics. Interestingly, BlueBERT$_{Inst.}$ outperformed other models on these two datasets.

**Supplementary Table S4**: Results of our models when evaluated using Nyflot and RO-ILS datasets. Each model was also fine-tuned 5 times and performance is reported as mean ± standard deviation.

| Model | Nyflot | | RO-ILS | |
|---|---|---|---|---|
| | F1 | ROC AUC | F1 | ROC AUC |
| SVM$_{Inst.}$ | 0.67 ± 0.00 | 0.67 ± 0.00 | 0.82 ± 0.00 | 0.92 ± 0.00 |
| BlueBERT$_{Inst.}$ | **0.82 ± 0.04** | **0.93 ± 0.04** | **0.96 ± 0.02** | **0.95 ± 0.04** |
| BlueBERT$_{TRANSFER}$ | 0.65 ± 0.19 | 0.91 ± 0.03 | 0.77 ± 0.20 | 0.62 ± 0.23 |